\documentclass{article}

\usepackage{arxiv}

\usepackage[utf8]{inputenc} 
\usepackage[T1]{fontenc}    

\usepackage{url}            
\usepackage{booktabs}       
\usepackage{amsfonts}       
\usepackage{nicefrac}       
\usepackage{microtype}      
\usepackage{lipsum}

\usepackage{amsmath}

\usepackage[colorlinks,linkcolor=red]{hyperref}

\usepackage{enumitem,amssymb}
\usepackage{graphicx}

\usepackage[justification=centering]{caption}
\usepackage{subcaption}
\usepackage{float}

\usepackage[ruled,vlined]{algorithm2e}
\usepackage{algorithmic}

\usepackage{listings}
\usepackage{xcolor}
\title{
Experimental Analysis of Legendre Decomposition in Machine Learning
}

\author{
  Jianye Pang\\
  Department of Computer Science\\
  Xi'an Jiaotong University\\
  \texttt{sherlockholmes@stu.xjtu.edu.cn}
  \And
  Kai Yi\\
  Department of Computer Science\\
  King Abdullah University of Science and Technology\\
  \texttt{kai.yi@kaust.edu.sa} \\
  \And
  Wanguang Yin\\
  Department of Computer Science\\
  Southern University of Science and Technology\\
  \texttt{yinwg@sustech.edu.cn} \\
  \And 
  Min Xu\thanks{Corresponding author}\\
  Computational Biology Department\\
  Carnegie Mellon University\\
  \texttt{mxu1@cs.cmu.edu}
}

\begin{document}
\maketitle

\begin{abstract}
In this technical report, we analyze Legendre decomposition for non-negative tensor in theory and application. In theory, the properties of dual parameters and dually flat manifold in Legendre decomposition are reviewed, and the process of tensor projection and parameter updating is analyzed. In application, a series of verification experiments and clustering experiments with parameters on submanifold were carried out, hoping to find an effective lower dimensional representation of the input tensor. The experimental results show that the parameters on submanifold have no ability to be directly used as low-rank representations. Combined with analysis, we connect Legendre decomposition with neural networks and low-rank representation applications, and put forward some promising prospects.
\end{abstract}  



%
\section{Introduction to Legendre Decomposition}
Matrix and tensor decomposition is the multiplication of a number of smaller matrices or tensors that are approximately disassembled by matrix and tensor. Up to now, the main matrix decomposition techniques have been widely used in computer vision, recommendation system, signal processing and other fields. Currently, standard methods for third-order non-negative tensor decomposition include CP decomposition\cite{anandkumar2014tensor} and Tucker decomposition\cite{kim2007nonnegative}.

It's well known the normal non-negative Tucker and CP tensor decomposition include non-convex optimization and that the global convergence is not guaranteed. One direction is to apply additional assumptions on data, such as a bounded variance, to transform the non-convex optimization problem into a convex one\cite{liu2012tensor, tomioka2013convex}. 

Legendre decomposition\cite{sugiyama2018legendre} is a new non-negative tensor decomposition method proposed by Mahito Sugiyama et al. Compared with the existing non-negative tensor decomposition methods, the greatest contribution of Legendre decomposition lies in the transformation of the non-convex optimization problem onto a convex submanifold space without additional assumptions, which ensures global convergence, and the use of gradient descent can find a unique reconstructed tensor satisfying and the minimum Kullback-Leibler (KL) divergence from the input matrix.

In this paper, we analyze Legendre tensor decomposition in both theory and application. From the perspective of theory, we aim to analyze the properties of dual parameters and dually flat manifold introduced in Legendre tensor decomposition. From the perspective of application, we aim to verify the ability of parameters on submanifold to represent the semantics of the input tensor and discuss whether there is a connection between Legendre decomposition technique and classical neural network structures and low-rank representations.

\section{Related Work}
\subsection{Method of Non-Negative CP and Tucker Decomposition}
The most fundamental methods in non-negative tensor decomposition lied on non-negative Tucker decomposition\cite{kim2007nonnegative} and non-negative CP decomposition\cite{anandkumar2014tensor}.

For non-negative Tucker decomposition. Given tensor $\mathcal{X}\in \mathbb{R}_{\geq 0}^{n_1\times n_2\times n_3}$, then $\mathcal{X} \approx \mathcal{G} \times_{1} U \times_{2} V \times_{3} W$, where $\mathcal{G}\in \mathbb{R}_{\geq 0}^{r_1\times r_2\times r_3}$ is the core tensor, $U\in \mathbb{R}_{\geq 0}^{n_1\times r_1}$, $V\in \mathbb{R}_{\geq 0}^{n_2\times r_2}$ and $W\in \mathbb{R}_{\geq 0}^{n_3\times r_3}$ are projection matrices. Thus we have

\begin{equation}
x_{i j k} \approx \sum_{m=1}^{r_{1}} \sum_{n=1}^{r_{2}} \sum_{l=1}^{r_{3}}\left(g_{m n l} \cdot u_{i m} \cdot v_{j n} \cdot w_{k l}\right).
\end{equation}

CP decomposition can be regarded as a special case of Tucker decomposition. The form of CP decomposition is $\mathcal{X} \approx \sum_{p=1}^{r} \lambda_{p} F(:, p) \circ S(:, p) \circ T(:, p)$, the core tensor $\lambda\in \mathbb{R}_{\geq 0}^{r\times r\times r}$. Also $F\in \mathbb{R}_{\geq 0}^{n_1\times r}$, $S\in \mathbb{R}_{\geq 0}^{n_2\times r}$ and $T\in \mathbb{R}_{\geq 0}^{n_3\times r}$ are factor matrices. We have

\begin{equation}
    x_{i j k} \approx \sum_{p=1}^{r} \lambda_{p} \cdot f_{i p} \cdot s_{j p} \cdot t_{k p}.
\end{equation}

\subsection{Dual Coordinates and Dually Flat manifold}
Two parameters that can be derived from each other by Legendre transformation and mapped one to one are called dual coordinates, which is described in detail in information geometry\cite{amari2008information}. For a convex function $\psi(\boldsymbol{\theta})$ of $\theta$, that satisfies:
$$\psi(\boldsymbol{\theta})=\boldsymbol{\theta} \cdot \boldsymbol{\eta}-\varphi(\boldsymbol{\eta})$$
$\eta$ and $\theta$ are a set of dual parameters, corresponding to the coordinates of the same point respectively, and each other can be obtained by Legendre transformation:
$$\boldsymbol{\eta}=\operatorname{Grad} \psi(\boldsymbol{\theta}), \boldsymbol{\theta}=\operatorname{Grad} \varphi(\boldsymbol{\eta})$$
Corresponding to the definition of the function of parameter $\theta$ in Legendre decomposition:
$$\psi(\theta)=\log \sum_{v \in \Omega} \exp \left(\sum_{u \in B} \zeta(u, v) \theta(u)\right), \zeta(u, v)=\left\{\begin{array}{ll}
1 & \text { if } u \leq v \\
0 & \text { otherwise }
\end{array}\right.$$
$\psi(\theta)$ is convex since its functions is member of the exponential family, the gradient is calculated by partial derivatives:
$$-\sum_{v \in \Omega} p_{v} \zeta(w, v)+\frac{\partial \psi(\theta)}{\partial \theta_{w}}=\eta_{w}-\hat{\eta}_{w}$$
Hence, $(\theta, \eta)$ is a set of dual parameters obtained by Legendre transformation.

A manifold with dual connections can be embed in a dually flat manifold of high dimensions with no limitations. Legendre decomposition builds dual coordinate system $(\theta,\eta)$ on normalized tensor $S$ from input tensor $X$, hence $S$ becomes a dually flat manifold. Similar to the Generalized Pythagorean Theorem in information geometry\cite{amari2008information}, consider replacing the points $P$, $Q$, and $R$ in $S$ with tensor to form the projection in statistical manifold. The set of all the discrete probability distributions which is transformed from tensor gives a dually flat manifold since any parameterized family of probability distributions over discrete random variables is a curved exponential family.

By proving that $D_{\mathrm{KL}}(\mathcal{P}, \mathcal{Q})$ is convex in section 2.2 from Legendre decomposition\cite{sugiyama2018legendre}, it proves that e-flat submanifold, e-projection, Legendre decomposition are all convex.

\subsection{Method of Legendre Decomposition}
Legendre decomposition\cite{sugiyama2018legendre} is realized as a projection of the input tensor onto a submanifold composed of reconstructable tensors. 

\subsubsection{Definition}
Given $\mathcal{X}\in \mathbb{R}_{\geq 0}^{I_1\times \cdots \times I_N}$ (if concerted to probabilisty mass function, then denotes $\mathcal{X}$ as $\mathcal{P}$), the sample space $\Omega \subset [I_1]\times \cdots \times [I_N]$ where $\left[I_{k}\right]=\left\{1,2, \ldots, I_{k}\right\}$, and a parameter basis $B\subset \Omega \setminus\{(1,1, \ldots, 1)\}$, Legendre decomposition finds the fully decomposable tensor $\mathcal{Q} \in \mathbb{R}_{\geq 0}^{I_{1} \times I_{2} \times \cdots \times I_{N}}$ with a $B$ that minimizes the KL divergence $D_{\mathrm{KL}}(\mathcal{P}, \mathcal{Q})=\sum_{v \in \Omega} p_{v} \log \left(p_{v} / q_{v}\right)$.

\subsubsection{Algorithm}
\begin{figure}[H]
\begin{center}
\includegraphics[width=0.8\linewidth]{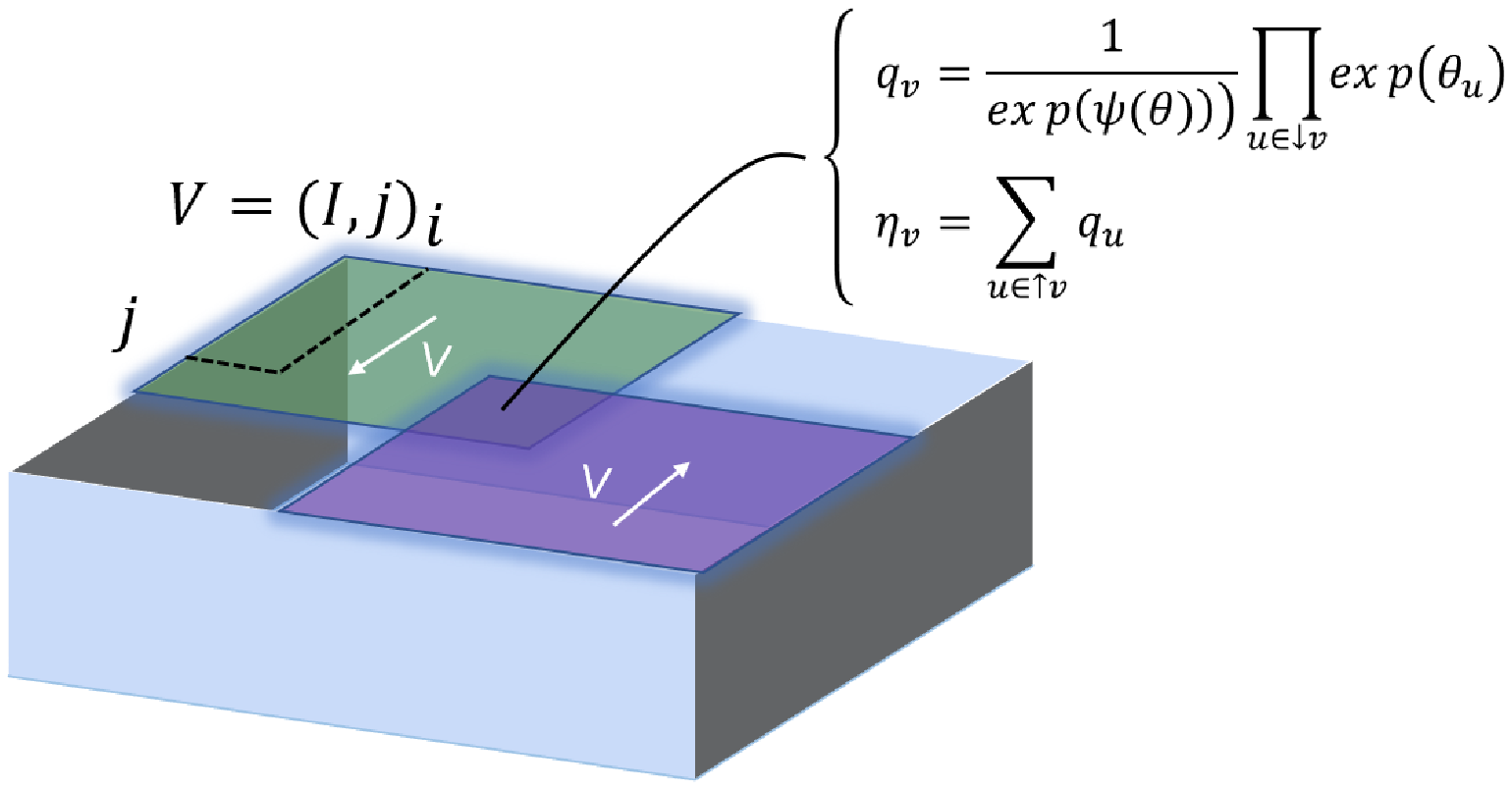}
\end{center}
\caption{Reconstruction process of Legendre decomposition.}
\label{rp}
\end{figure}

Legendres decomposition normalizes any non-negative tensor and transforms it into a set of discrete probability distributions with partial order in a statistical manifold. By introducing Legendre dual parameters in information geometry given in Figure~\ref{rp}, the mapping of the tensor onto dually flat manifold is achieved. The tensor decomposition task on submanifold is the updating and optimization process of parameters and basis. By minimizing the KL divergence between the input tensor and the reconstructed tensor, it guarantees that the Legendre decomposition is convex.\\
Computing parameters greater than and less than $v$ in the figure is similar to information decomposition\cite{sugiyama2016information} upward and downward in the itemset lattice of two patterns, which corresponding to pattern mining and log-linear analysis respectively. A similar definition of parameter is also applied to tensor balancing\cite{sugiyama2017tensor} to prove its duality.  

\section{Experiments for Validation} \label{sec4}
\subsection{Datasets}
Motivated from Legendre decomposition\cite{sugiyama2018legendre}, two image datasets MNIST\cite{lecun2010mnist} and face image dataset\footnote{This dataset is originally distributed at http://www.cl.cam.ac.uk/research/dtg/attarchive/
facedatabase.html and also available from the R rTensor package (https://CRAN.R-project.org/
package=rTensor.} are used to generate 3D tensors of size 28*28*500. In this paper, MNIST is chosen to conduct the experiments. For tensor decomposition from digits of 0-9, 100 images of each digit were used to splicing into 10 28*28*100 tensors, while other experiments used 28*28*100 tensors generated by the digit 8. All the experiments in Section~\ref{sec4} except partial sort analysis were done without partial order limitation on $\Omega$ and were based on natural gradient. The basic properties of Legendre decomposition are analyzed mainly through experiments for validation. Source code could refer to \href{https://github.com/sherjy/LegendreDecomposition}{https://github.com/sherjy/LegendreDecomposition}.

\subsection{Evaluation Metrics}
Following metrics are used to evaluate compared methods of experiments.\\
\textbf{RMSE} evaluates the quality of decomposition by the root mean squared error (RMSE) between the input and the reconstructed tensors.\\
\textbf{Running time} is calculated as the average time from a batch of tensors, the batch is designed manually (e.g. 100).

\subsection{Arguments Recap}
Recap key arguments used in Legendre decomposition program here to restate the detailed settings for each experiment. The main arguments used include $-c, -n, -d, -b, -i$.
\begin{itemize}
    \item Argument $-c(core size)$ means the parameter for a decomposition basis. Hence $N_{par}$ equals to $c \cdot N_3$ if the tensor with partial order is three-dimensional.
    \item Argument $-n$ means the natural gradient is used.
    \item Argument $-d$ means the depth size, which is also equal to the last dimension of the input tensor.
    \item Argument $-b$ means type of a decomposition basis (1 means w/o partial order (random), 2 means partial order on $\beta$). 
    \item Argument $-i$ means input file from test0.csv to test9.csv (digits 0-9).
\end{itemize}

\subsection{Decomposition for Each Digits}
\begin{table}[!htb]
\centering
\caption{RMSE statistical experiment (w/o sort)}
\label{RMSE statistical}
\begin{tabular}{ccccc}
\toprule
Digit & $N_{par}$ & $N_{iter}$ & running time & RMSE    \\ \hline
0        & 5054                 & 6                    & 295.851      & 31.8912 \\
1        & 5054                 & 6                    & 295.229      & 14.6909 \\
2        & 5054                 & 5                    & 270.418      & 30.3843 \\
3        & 5054                 & 6                    & 317.698      & 29.5802 \\
4        & 5054                 & 6                    & 479.203      & 26.7226 \\
5        & 5054                 & 6                    & 489.082      & 27.9886 \\
6        & 5054                 & 6                    & 342.573      & 28.289  \\
7        & 5054                 & 6                    & 336.42       & 23.5918 \\
8        & 5054                 & 6                    & 318.538      & 30.4651 \\
9        & 5054                 & 6                    & 304.977      & 25.8854 \\ \bottomrule
\end{tabular}
\end{table}

\textbf{RMSE statistical experiment settings:}
-c:50
-n:natural gradient
-d:100
-b:1
-i:from test0.csv to test9.csv
theta:0\\
\textbf{Analysis:} Legendre decomposition statistics of digits 0-9 are given in Table~\ref{RMSE statistical}. From the table, the digit 1 has the lowest decomposition RMSE, which is consistent with the intuition that the digit 1 is the simplest and easiest to write. The digit 0,1 and 2 has the minimum running time. Since during the decomposition, most of the time of decomposition is saved because the pixel value of most positions in the image matrix of digit 1 is 0.

\subsection{Initialization Forms}
To demonstrate the impact of parameter $\left(\theta_{v}\right)_{v \in B}$ initialization on tensor reconstruction, we compare all-zero initialization (paper) with the random initialization, the uniform initialization and the gaussian initialization. 

\begin{table*}[!htb]
\begin{minipage}{\textwidth}
 \begin{minipage}[t]{0.45\textwidth}
  \centering
     \makeatletter\def\@captype{table}\makeatother\caption{All-zero initialization experiment (w/o sort)}
   \label{w/o}
     \resizebox{\textwidth}{18mm}{
      \begin{tabular}{lllll}
\hline
variable & $N_{par}$ & $N_{iter}$ & running time & RMSE    \\ \hline
15       & 1554                 & 5                    & 7.02333      & 36.198  \\
20       & 2054                 & 5                    & 16.164       & 35.3355 \\
25       & 2554                 & 5                    & 31.3414      & 34.4837 \\
30       & 3054                 & 5                    & 54.05        & 33.673  \\
35       & 3554                 & 5                    & 85.9133      & 32.9019 \\
40       & 4054                 & 5                    & 130.025      & 31.997  \\
45       & 4554                 & 6                    & 221.035      & 31.2293 \\
50       & 5054                 & 6                    & 296.163      & 30.4651 \\ \hline
\end{tabular}
}
  \end{minipage}
  \hspace{10pt}
  \begin{minipage}[t]{0.45\textwidth}
   \centering
        \makeatletter\def\@captype{table}\makeatother\caption{Random distribution experiment (w/o sort)}
\label{Random}
\resizebox{\textwidth}{18mm}{
\begin{tabular}{lllll}
\hline
variable & $N_{par}$ & $N_{iter}$ & running time & RMSE    \\ \hline
15       & 1554                 & 5                    & 7.11238      & 36.198  \\
20       & 2054                 & 5                    & 17.4566      & 35.3355 \\
25       & 2554                 & 5                    & 32.2854      & 34.4837 \\
30       & 3054                 & 5                    & 58.965       & 33.673  \\
35       & 3554                 & 5                    & 88.913       & 32.9019 \\
40       & 4054                 & 5                    & 131.812      & 31.997  \\
45       & 4554                 & 6                    & 220.716      & 31.2293 \\
50       & 5054                 & 6                    & 302.211      & 30.4651 \\ \hline
\end{tabular}
}
   \end{minipage}
\end{minipage}
\end{table*}

\begin{table*}[!htb]
\begin{minipage}{\textwidth}
 \begin{minipage}[t]{0.45\textwidth}
  \centering
     \makeatletter\def\@captype{table}\makeatother\caption{Uniform distribution experiment (w/o sort)}
\label{Uniform}
     \resizebox{\textwidth}{18mm}{
\begin{tabular}{lllll}
\hline
variable & $N_{par}$ & $N_{iter}$ & running time & RMSE    \\ \hline
15       & 1554                 & 5                    & 6.96107      & 36.198  \\
20       & 2054                 & 5                    & 16.7065      & 35.3355 \\
25       & 2554                 & 5                    & 32.1848      & 34.4837 \\
30       & 3054                 & 5                    & 54.7099      & 33.673  \\
35       & 3554                 & 5                    & 86.8046      & 32.9019 \\
40       & 4054                 & 5                    & 136.108      & 31.997  \\
45       & 4554                 & 6                    & 229.096      & 31.2293 \\
50       & 5054                 & 6                    & 299.731      & 30.4651 \\ \hline
\end{tabular}}
  \end{minipage}
    \hspace{10pt}
  \begin{minipage}[t]{0.45\textwidth}
   \centering
        \makeatletter\def\@captype{table}\makeatother\caption{Gaussian distribution experiment (w/o sort)}
\label{Gaussian}
\resizebox{\textwidth}{18mm}{
\begin{tabular}{lllll}
\hline
variable & $N_{par}$ & $N_{iter}$ & running time & RMSE    \\ \hline
15       & 1554                 & 5                    & 7.10355      & 36.198  \\
20       & 2054                 & 5                    & 17.0781      & 35.3355 \\
25       & 2554                 & 5                    & 35.0911      & 34.4837 \\
30       & 3054                 & 5                    & 55.6617      & 33.673  \\
35       & 3554                 & 5                    & 88.3894      & 32.9019 \\
40       & 4054                 & 5                    & 130.677      & 31.997  \\
45       & 4554                 & 6                    & 218.843      & 31.2293 \\
50       & 5054                 & 6                    & 304.596      & 30.4651 \\ \hline
\end{tabular}
}
   \end{minipage}
\end{minipage}
\end{table*}

\textbf{All-zero initialization experiment settings:}
-c:15, 20, 25, 30, 35, 40, 45, 50
-n:natural gradient
-d:100
-b:1
-i:test8.csv
theta:0

\textbf{Random distribution experiment settings:}
-c:15, 20, 25, 30, 35, 40, 45, 50
-n:natural gradient
-d:100
-b:1
-i:test8.csv
theta:random (0, max(theta))


\textbf{Uniform distribution experiment settings:}
-c:15, 20, 25, 30, 35, 40, 45, 50
-n:natural gradient
-d:100
-b:1
-i:test8.csv
theta:(0, max(theta)) divided by $n1 \cdot n2 \cdot n3$


\textbf{Gaussian distribution experiment settings:}
-c:15, 20, 25, 30, 35, 40, 45, 50
-n:natural gradient
-d:100
-b:1
-i:test8.csv
theta:$$\frac{1}{\sqrt{2 \pi}} e^{-\frac{x^{2}}{2}}, \text { where } x=(i-1) \cdot n_2 \cdot n_3+(j-1) \cdot n_3+k-\frac{n_1 \cdot n_{2} \cdot n_3}{2}$$

\textbf{Analysis:} Compared with all-zero initialization for $\theta$ from Table~\ref{w/o}, three contrastive initializations on tensor reconstruction are given in Table~\ref{Random}, Table~\ref{Uniform} and Table~\ref{Gaussian}. From the tables, different initializations have little impact on running time, and the RMSE of the reconstructed tensor is the same. As the parameters for a decomposition basis used increase, the running time and number of iterations keep going up, and RMSE keeps going down.



\subsection{Partial Sort on Decomposition Basis}
To demonstrate the impact of partial sort on decomposition basis $\beta$, we enable the w/ partial order mode in the current experiment and compare it with the w/o partial order (random) mode we have been using in the previous experiments.

\begin{table}[!htb]
\centering
\caption{W/ sort experiment}
\label{w/}
\begin{tabular}{lllll}
\hline
variable & $N_{par}$ & $N_{iter}$ & running time & RMSE    \\ \hline
15       & 1500                 & 4                    & 5.17718      & 36.6394 \\
20       & 2000                 & 4                    & 12.2843      & 35.8749 \\
25       & 2500                 & 4                    & 23.7491      & 35.2418 \\
30       & 3000                 & 4                    & 40.9128      & 34.5281 \\
35       & 3500                 & 4                    & 64.934       & 33.8396 \\
40       & 4000                 & 4                    & 100.498      & 33.246  \\
45       & 4500                 & 4                    & 140.251      & 32.5536 \\
50       & 5000                 & 4                    & 188.182      & 31.7547 \\ \hline
\end{tabular}
\end{table}
\textbf{W/ sort experiment settings:}
-c:15, 20, 25, 30, 35, 40, 45, 50
-n:natural gradient
-d:100
-b:2
-i:test8.csv
theta:0\\
\textbf{Analysis:} The two modes random order and partial order on decomposition basis are given in Table~\ref{w/o} and Table~\ref{w/}. \emph{From the tables, there are doubts that why w/o sort works better than w/ sort and the random mode can restore the original tensor. In theory, the problem is convex and the reconstructed tensor is optimal only in w/ sort mode, which conflict with the result of using the random mode in the code. For example, the RMSE results from Table~\ref{w/o} to Table~\ref{Gaussian} indicate that the reconstruction of the tensor is unique for w/o partial sort, whereas it should be unique for w/ partial order inspired from Legendre decomposition\cite{sugiyama2018legendre}. We have not yet been able to explain why this happens.}

\section{Experiments for Investigation} \label{Investigation}
\subsection{Datasets}
The following experiments of parameter clustering were conducted on 28x28x10/28x28 tensor datasets using digits 0-9 generated by MNIST. Each digit has 100 tensors, so there are 1,000 of them in total. The purpose of the investigation experiments is to verify that if the parameters during Legendre decomposition can perform rank reduction to represent the input tensor as low-rank representation through parameters clustering. Use \href{https://cran.r-project.org/web/packages/rTensor/index.html}{rTensor} tools to generate tensors. \\

\subsection{Evaluation Metrics}
This paper chooses two commonly used clustering evaluation indicators: adjusting mutual information (AMI), and adjusting random index (ARI) on clustering results.\\
Rand index (RI) is to measure the similarity of two cluster classes. Assuming that the number of samples is N and $C_{n}^{2}$ is the number of all possible sample pairs, which is defined as follows:
$$R I=\frac{a+b}{C_{n}^{2}}$$
\textbf{ARI} solves the problem that RI cannot well describe the similarity of randomly assigned cluster class marker vectors, which is defined as follows:
$$A R I=\frac{R I-E[R I]}{\max (R I)-E[R I]}$$
\textbf{AMI} is based on the mutual information score of predicted cluster vectors and real cluster vectors to measure their similarity, which is defined as follows:
$$A M I(U, V)=\frac{M I(U, V)-E\{M I(U, V)\}}{\max \{H(U), H(V)\}-E\{M I(U, V)\}}$$

\subsection{Parameters Clarification}
The original input tensor is $X$, which corresponds to $S$ in the submanifold space. Each element of the tensor $S$ is $s$ and has four main attributes, $p$, $\theta$, $\sum{\theta}$, and $N_{noneZero}$.
\begin{itemize}
    \item The attribute $p$ holds information about $s$ restoring $X$.
    $$\mathcal{P}=\mathcal{X} / \sum_{v} x_{v}$$
    \item The attributes $\theta$ and $\eta$ are iteration parameters. They are used in the reconstruction process of $q$ value, and $p$ is used instead of $q$ in the state of e-projection.
    $$\log q_{v}=\sum_{u \in \Omega^{+}} \zeta(u, v) \theta_{u}-\psi(\theta)=\sum_{u \in \Omega} \zeta(u, v) \theta_{u}, \quad \zeta(u, v)=\left\{\begin{array}{ll}
1 & \text { if } u \leq v \\
0 & \text { otherwise }
\end{array}\right.$$
    \item The attribute $\sum{\theta}$ is a sum of all $\theta$ in tensor $S$.
    \item The attribute $N_{noneZero}$ is a count of $p$ values that are not zero in tensor $S$.
    \item In addition, $\beta$ stores basis positions and a series of $\hat{\eta}$ values in the calculation. The number of $\beta$ parameters is equal to core\_size times the last dimension of the input tensor. That is, $\beta$ slices the last dimension of $X$, and selects core\_size basis on each slice.
    \item $D_{KL}$ is the target optimization with $\theta$ and $\eta$.
    $$\mathcal{Q}=\underset{\mathcal{R} \in \mathcal{S}_{B}}{\operatorname{argmin}} D_{\mathrm{KL}}(\mathcal{P}, \mathcal{R})$$
\end{itemize}
The dimensions of all attributes in $S$ are the same as those of the original input tensor $X$. \\

\subsection{Clustering of Image-batch Tensors} \label{image-batch}
\subsubsection{Parameters Clustering Settings} The following clustering experiments were conducted on 28x28x10 tensors, k-means is mainly used to verify the clustering experiments. The core\_size is set to 50 and cluster is set to 10. The purpose of the experiment~\ref{image-batch} is to verify whether the statistical characteristic on image-batch tensors has the property of low-rank representation.

\subsubsection{Clustering with Different Parameters} \label{exp1g}
For convenience, the following tables describe Digit by $D$ and Class by $C$.
\begin{table*}[!htb]
\begin{minipage}{\textwidth}
 \begin{minipage}[t]{0.45\textwidth}
  \centering
     \makeatletter\def\@captype{table}\makeatother\caption{Clustering of $\sum{\theta}$ and $\sum{\eta}$}
     \resizebox{\textwidth}{20mm}{
      \begin{tabular}{lllllllllll}
\hline
       & D0 & D1 & D2 & D3 & D4 & D5 & D6 & D7 & D8 & D9 \\\hline
C0 & 9      & 6      & 13     & 13     & 9      & 16     & 7      & 14     & 5      & 15     \\
C1 & 4      & 2      & 14     & 5      & 13     & 6      & 10     & 11     & 9      & 11     \\
C2 & 19     & 22     & 13     & 14     & 9      & 16     & 12     & 10     & 14     & 13     \\
C3 & 9      & 11     & 8      & 5      & 4      & 4      & 1      & 4      & 10     & 5      \\
C4 & 19     & 14     & 9      & 21     & 17     & 15     & 18     & 19     & 20     & 13     \\
C5 & 0      & 3      & 0      & 3      & 4      & 4      & 3      & 6      & 2      & 4      \\
C6 & 15     & 12     & 14     & 13     & 17     & 18     & 15     & 6      & 19     & 12     \\
C7 & 15     & 14     & 14     & 7      & 7      & 8      & 14     & 12     & 11     & 10     \\
C8 & 1      & 4      & 0      & 0      & 0      & 0      & 0      & 0      & 0      & 0      \\
C9 & 9      & 12     & 15     & 19     & 20     & 13     & 20     & 18     & 10     & 17    \\
\hline
\end{tabular}}
The vector shape of the cluster is [1x2,1].
  \end{minipage}
    \hspace{10pt}
  \begin{minipage}[t]{0.45\textwidth}
   \centering
        \makeatletter\def\@captype{table}\makeatother\caption{Clustering of $\sum{\theta}$ and $\sum{\eta}$ in $\beta$}
        \resizebox{\textwidth}{20mm}{
        \begin{tabular}{lllllllllll}
\hline
       & D0 & D1 & D2 & D3 & D4 & D5 & D6 & D7 & D8 & D9 \\\hline
C0 & 30     & 4      & 29     & 21     & 17     & 17     & 19     & 16     & 15     & 12     \\
C1 & 3      & 1      & 10     & 21     & 30     & 19     & 33     & 10     & 11     & 22     \\
C2 & 0      & 38     & 0      & 0      & 0      & 0      & 0      & 0      & 1      & 0      \\
C3 & 11     & 28     & 16     & 2      & 0      & 2      & 0      & 9      & 20     & 1      \\
C4 & 0      & 2      & 2      & 9      & 14     & 7      & 6      & 10     & 2      & 22     \\
C5 & 15     & 12     & 20     & 9      & 3      & 12     & 3      & 4      & 19     & 6      \\
C6 & 9      & 3      & 22     & 19     & 24     & 23     & 30     & 16     & 19     & 21     \\
C7 & 0      & 2      & 0      & 3      & 4      & 0      & 0      & 4      & 1      & 3      \\
C8 & 26     & 10     & 1      & 6      & 3      & 11     & 5      & 13     & 11     & 5      \\
C9 & 6      & 0      & 0      & 10     & 5      & 9      & 4      & 18     & 1      & 8     \\\hline
\end{tabular}}
The vector shape of the cluster is [1x2,1].
   \end{minipage}
\end{minipage}
\end{table*}

\begin{table*}[!htb]
\begin{minipage}{\textwidth}
 \begin{minipage}[t]{0.45\textwidth}
  \centering
     \makeatletter\def\@captype{table}\makeatother\caption{Clustering of $\sum{\theta}$ and $N_{noneZero}$}
     \resizebox{\textwidth}{20mm}{
      \begin{tabular}{lllllllllll}
\hline
       & D0 & D1 & D2 & D3 & D4 & D5 & D6 & D7 & D8 & D9 \\\hline
C0 & 8      & 6      & 23     & 11     & 31     & 5      & 5      & 5      & 11     & 13     \\
C1 & 14     & 4      & 21     & 21     & 15     & 23     & 17     & 20     & 18     & 15     \\
C2 & 24     & 15     & 2      & 12     & 13     & 19     & 10     & 14     & 10     & 9      \\
C3 & 19     & 13     & 2      & 6      & 3      & 7      & 2      & 21     & 13     & 10     \\
C4 & 7      & 13     & 14     & 10     & 12     & 9      & 21     & 9      & 15     & 17     \\
C5 & 5      & 22     & 3      & 8      & 3      & 9      & 8      & 3      & 8      & 2      \\
C6 & 0      & 12     & 0      & 0      & 0      & 0      & 0      & 5      & 1      & 0      \\
C7 & 15     & 5      & 21     & 24     & 16     & 27     & 33     & 15     & 19     & 22     \\
C8 & 2      & 9      & 0      & 0      & 0      & 0      & 0      & 0      & 0      & 0      \\
C9 & 6      & 1      & 14     & 8      & 7      & 1      & 4      & 8      & 5      & 12    \\\hline
\end{tabular}}
The vector shape of the cluster is [1x2,1].
  \end{minipage}
    \hspace{10pt}
  \begin{minipage}[t]{0.45\textwidth}
   \centering
        \makeatletter\def\@captype{table}\makeatother\caption{Clustering of $\sum{p}$ and $N_{noneZero}$}
        \resizebox{\textwidth}{20mm}{
        \begin{tabular}{lllllllllll}
\hline
       & D0 & D1 & D2 & D3 & D4 & D5 & D6 & D7 & D8 & D9 \\\hline
C0 & 45     & 0      & 11     & 13     & 2      & 5      & 9      & 0      & 24     & 0      \\
C1 & 0      & 0      & 0      & 2      & 29     & 6      & 7      & 33     & 0      & 29     \\
C2 & 15     & 0      & 29     & 26     & 1      & 12     & 11     & 0      & 37     & 0      \\
C3 & 0      & 70     & 0      & 0      & 0      & 0      & 0      & 0      & 0      & 0      \\
C4 & 0      & 2      & 0      & 0      & 11     & 3      & 1      & 39     & 0      & 6      \\
C5 & 4      & 0      & 35     & 26     & 4      & 13     & 25     & 4      & 27     & 15     \\
C6 & 0      & 0      & 6      & 15     & 30     & 34     & 21     & 19     & 2      & 31     \\
C7 & 1      & 0      & 17     & 18     & 23     & 27     & 26     & 5      & 7      & 19     \\
C8 & 0      & 28     & 0      & 0      & 0      & 0      & 0      & 0      & 0      & 0      \\
C9 & 35     & 0      & 2      & 0      & 0      & 0      & 0      & 0      & 3      & 0     \\\hline
\end{tabular}}
The vector shape of the cluster is [1x2,1].
   \end{minipage}
\end{minipage}
\end{table*}

\begin{table*}[!htb]
\begin{minipage}{\textwidth}
 \begin{minipage}[t]{0.45\textwidth}
  \centering
      \makeatletter\def\@captype{table}\makeatother\caption{Clustering of last $D_{KL}$}
     \resizebox{\textwidth}{20mm}{
      \begin{tabular}{lllllllllll}
\hline
       & D0 & D1 & D2 & D3 & D4 & D5 & D6 & D7 & D8 & D9 \\\hline
C0 & 5      & 0      & 8      & 5      & 1      & 2      & 3      & 0      & 3      & 1      \\
C1 & 0      & 6      & 0      & 0      & 0      & 0      & 0      & 1      & 0      & 0      \\
C2 & 0      & 36     & 0      & 0      & 1      & 2      & 0      & 10     & 0      & 4      \\
C3 & 1      & 17     & 2      & 3      & 9      & 10     & 2      & 17     & 2      & 6      \\
C4 & 0      & 30     & 0      & 0      & 1      & 0      & 1      & 5      & 1      & 0      \\
C5 & 22     & 0      & 24     & 25     & 20     & 23     & 22     & 17     & 23     & 27     \\
C6 & 25     & 0      & 25     & 24     & 9      & 11     & 29     & 9      & 27     & 18     \\
C7 & 7      & 2      & 18     & 25     & 37     & 28     & 15     & 17     & 19     & 15     \\
C8 & 35     & 0      & 18     & 10     & 6      & 7      & 17     & 3      & 15     & 9      \\
C9 & 5      & 9      & 5      & 8      & 16     & 17     & 11     & 21     & 10     & 20    \\\hline
\end{tabular}}
The vector shape of the cluster is [1,1].
  \end{minipage}
    \hspace{10pt}
  \begin{minipage}[t]{0.45\textwidth}
   \centering
        \makeatletter\def\@captype{table}\makeatother\caption{The performance of the statistical characteristic clustering on image-batch tensors}
        \label{res4.4.2}
        \begin{tabular}{lll}
\hline
Clustering & AMI             & ARI              \\ \hline
$\sum{\theta}$ and $\sum{\eta}$      & 0.01017  & 0.00285 \\
$\sum{\theta}$ and $\sum{\eta}$ in $\beta$      & 0.12785  & 0.06186  \\
$\sum{\theta}$ and $N_{noneZero}$      & 0.05209 & 0.01933  \\
$\sum{p}$ and $N_{noneZero}$      & 0.33020  & 0.19763   \\
$D_{KL}$ & 0.12238  & 0.04798    \\ \hline
\end{tabular}\\
The 2D visualization results of the original parameters and the statistical characteristic clustering experiments are shown in Figure~\ref{Original2.7.1} to Figure~\ref{Original2.7.4} and Figure~\ref{Predicted2.7.1} to Figure~\ref{Predicted2.7.4} in appendix.
   \end{minipage}
\end{minipage}
\end{table*}

\subsubsection{Analysis}
\begin{figure}[H]
\begin{subfigure}{.5\textwidth}
  \centering
\includegraphics[width=0.9\linewidth, height=0.8\textwidth, trim={20 20 20 20}, clip]{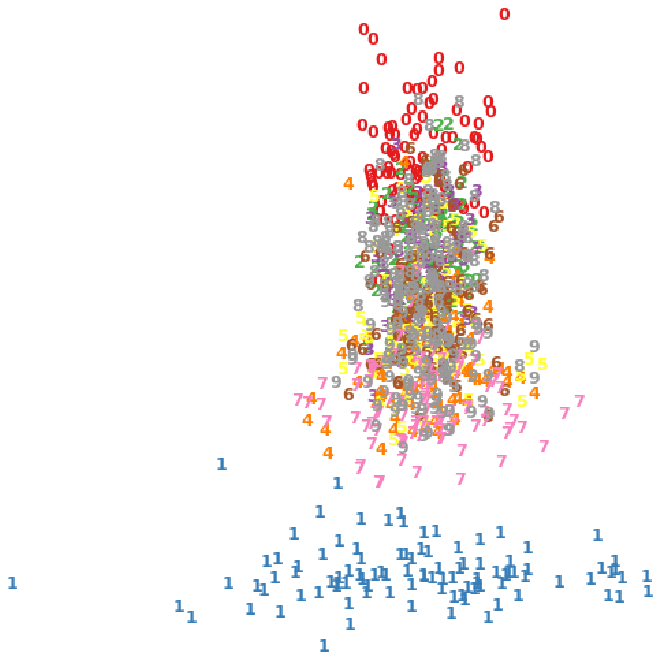}
  \caption{Original of $\sum{p}$ and $N_{noneZero}$}
  \label{Original2.7.4}
\end{subfigure}
\begin{subfigure}{.5\textwidth}
  \centering
\includegraphics[width=0.9\linewidth, height=0.8\textwidth, trim={20 20 20 20}, clip]{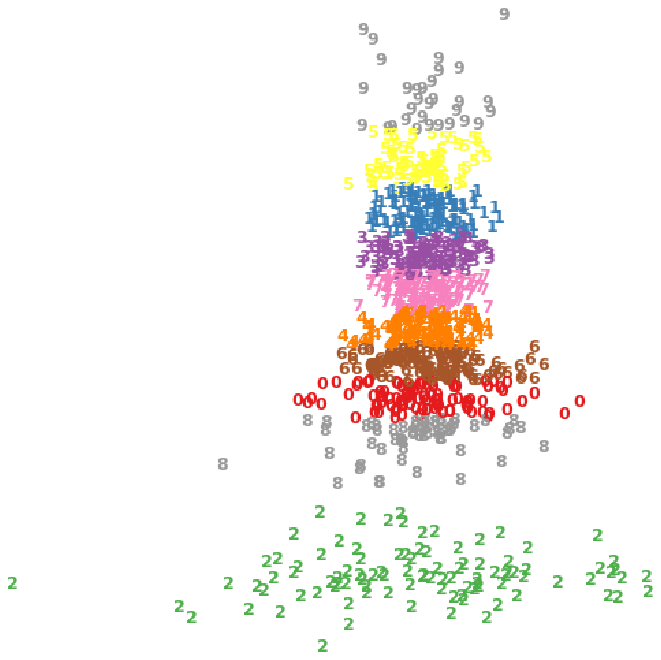}
  \caption{Clustering of $\sum{p}$ and $N_{noneZero}$}
  \label{Predicted2.7.4}
\end{subfigure}
\end{figure}

Table~\ref{res4.4.2} shows the results of AMI and ARI in each of experiments in Section~\ref{exp1g}. The value range is [-1,1], which is optimal when 1 is taken. Judging from the results of the evaluation indicators, the clustering effect of using statistical characteristics such as sum and count on image-batch is relatively poor. The ARI in all groups of experiments is close to 0, hence the clustering result is similar to the random generation. The AMI in the first three groups and the last group is only about 0.1, indicating that the clustering effect is extremely inconsistent with the real situation. As can be seen from Figure~\ref{Original2.7.1} to Figure~\ref{Original2.7.4}, the parameter is almost chaotic and non-separable from the original distribution, which indicates the statistical characteristics on image-batch tensors don't own the property of low-rank representation.

\subsection{Clustering of Single Image Unfolded tensor} \label{unfolding}
\subsubsection{Parameters Clustering Settings} The following clustering experiments were conducted on 28x28 tensors, k-means is mainly used to verify the clustering experiments. The core\_size is set to 25 and cluster is set to 10. The purpose of the experiment~\ref{unfolding} is to verify whether the unfolded feature on single image tensor has the property of low-rank representation.
\subsubsection{Clustering with Different Parameters}
\begin{table*}[!htb]
\begin{minipage}{\textwidth}
 \begin{minipage}[t]{0.45\textwidth}
  \centering
     \makeatletter\def\@captype{table}\makeatother\caption{Clustering of $p$}
     \label{table13}
     \resizebox{\textwidth}{20mm}{
      \begin{tabular}{lllllllllll}
\hline
       & D0 & D1 & D2 & D3 & D4 & D5 & D6 & D7 & D8 & D9 \\\hline
C0 & 0      & 0      & 1      & 3      & 55     & 3      & 0      & 17     & 4      & 60     \\
C1 & 4      & 0      & 4      & 89     & 0      & 47     & 0      & 1      & 15     & 1      \\
C2 & 0      & 0      & 0      & 0      & 23     & 0      & 0      & 73     & 1      & 33     \\
C3 & 1      & 2      & 84     & 1      & 2      & 0      & 3      & 1      & 2      & 0      \\
C4 & 0      & 57     & 1      & 0      & 5      & 0      & 1      & 0      & 1      & 0      \\
C5 & 0      & 38     & 0      & 0      & 0      & 0      & 0      & 0      & 1      & 0      \\
C6 & 0      & 3      & 6      & 6      & 6      & 8      & 1      & 3      & 76     & 3      \\
C7 & 0      & 0      & 0      & 0      & 0      & 32     & 0      & 0      & 0      & 0      \\
C8 & 92     & 0      & 0      & 0      & 5      & 5      & 6      & 5      & 0      & 3      \\
C9 & 3      & 0      & 4      & 1      & 4      & 5      & 89     & 0      & 0      & 0     \\\hline
\end{tabular}}
The vector shape of the cluster is [28x28,1].
  \end{minipage}
    \hspace{10pt}
  \begin{minipage}[t]{0.45\textwidth}
   \centering
        \makeatletter\def\@captype{table}\makeatother\caption{Clustering of $\theta$ and $\eta$}
        \label{table14}
        \resizebox{\textwidth}{20mm}{
        \begin{tabular}{lllllllllll}
\hline
       & D0 & D1 & D2 & D3 & D4 & D5 & D6 & D7 & D8 & D9 \\\hline
C0 & 0      & 35     & 1      & 3      & 6      & 4      & 10     & 6      & 6      & 9      \\
C1 & 0      & 0      & 0      & 0      & 2      & 3      & 0      & 18     & 0      & 8      \\
C2 & 13     & 2      & 21     & 8      & 4      & 14     & 8      & 8      & 8      & 13     \\
C3 & 76     & 0      & 44     & 38     & 10     & 23     & 18     & 4      & 15     & 6      \\
C4 & 4      & 0      & 16     & 4      & 35     & 10     & 52     & 3      & 4      & 13     \\
C5 & 4      & 0      & 9      & 43     & 12     & 23     & 0      & 29     & 43     & 22     \\
C6 & 1      & 43     & 7      & 4      & 15     & 8      & 11     & 12     & 11     & 17     \\
C7 & 2      & 5      & 2      & 0      & 16     & 15     & 1      & 17     & 13     & 12     \\
C8 & 0      & 10     & 0      & 0      & 0      & 0      & 0      & 3      & 0      & 0      \\
C9 & 0      & 5      & 0      & 0      & 0      & 0      & 0      & 0      & 0      & 0     \\\hline
\end{tabular}}
The vector shape of the cluster is [28x28x2,1].
   \end{minipage}
\end{minipage}
\end{table*}

\begin{table*}[!htb]
\begin{minipage}{\textwidth}
 \begin{minipage}[t]{0.45\textwidth}
  \centering
     \makeatletter\def\@captype{table}\makeatother\caption{Clustering of $\theta$ and $\eta$ in $\beta$}
     \label{table15}
     \resizebox{\textwidth}{20mm}{
      \begin{tabular}{lllllllllll}
\hline
       & D0 & D1 & D2 & D3 & D4 & D5 & D6 & D7 & D8 & D9 \\\hline
C0 & 1      & 11     & 1      & 3      & 3      & 1      & 2      & 6      & 2      & 2      \\
C1 & 21     & 14     & 18     & 20     & 24     & 16     & 17     & 18     & 18     & 15     \\
C2 & 0      & 0      & 0      & 0      & 0      & 0      & 1      & 2      & 0      & 1      \\
C3 & 10     & 10     & 9      & 12     & 14     & 16     & 7      & 8      & 13     & 16     \\
C4 & 39     & 27     & 30     & 31     & 18     & 19     & 25     & 30     & 39     & 27     \\
C5 & 0      & 10     & 3      & 0      & 2      & 1      & 2      & 4      & 1      & 3      \\
C6 & 2      & 5      & 8      & 2      & 9      & 13     & 9      & 5      & 3      & 9      \\
C7 & 27     & 21     & 31     & 32     & 30     & 34     & 37     & 26     & 24     & 27     \\
C8 & 0      & 0      & 0      & 0      & 0      & 0      & 0      & 1      & 0      & 0      \\
C9 & 0      & 2      & 0      & 0      & 0      & 0      & 0      & 0      & 0      & 0     \\\hline
\end{tabular}}
The vector shape of the cluster is [50x2,1].
  \end{minipage}
    \hspace{10pt}
  \begin{minipage}[t]{0.45\textwidth}
   \centering
        \makeatletter\def\@captype{table}\makeatother\caption{The performance of the unfolded feature clustering on single image tensor}
        \label{res4.4.3}
        \begin{tabular}{lll}
\hline
Clustering & AMI             & ARI              \\ \hline
$p$      & 0.50459  & 0.36869   \\
$\theta$ and $\eta$      & 0.46534  & 0.33989   \\
$\theta$ and $\eta$ in $\beta$      & 0.01402 & 0.00648 \\ \hline
\end{tabular} \\
The t-SNE visualization results of the original parameters and the unfolded feature clustering experiments are shown in Figure~\ref{Original2.8.1} to Figure~\ref{Original2.8.3} and Figure~\ref{Predicted2.8.1} to Figure~\ref{Predicted2.8.3} in appendix.
   \end{minipage}
\end{minipage}
\end{table*}

\subsubsection{Analysis}
\begin{figure}[!htb]
\begin{subfigure}{.5\textwidth}
  \centering
  \includegraphics[width=0.9\linewidth, height=0.8\textwidth, trim={20 20 20 20}, clip]{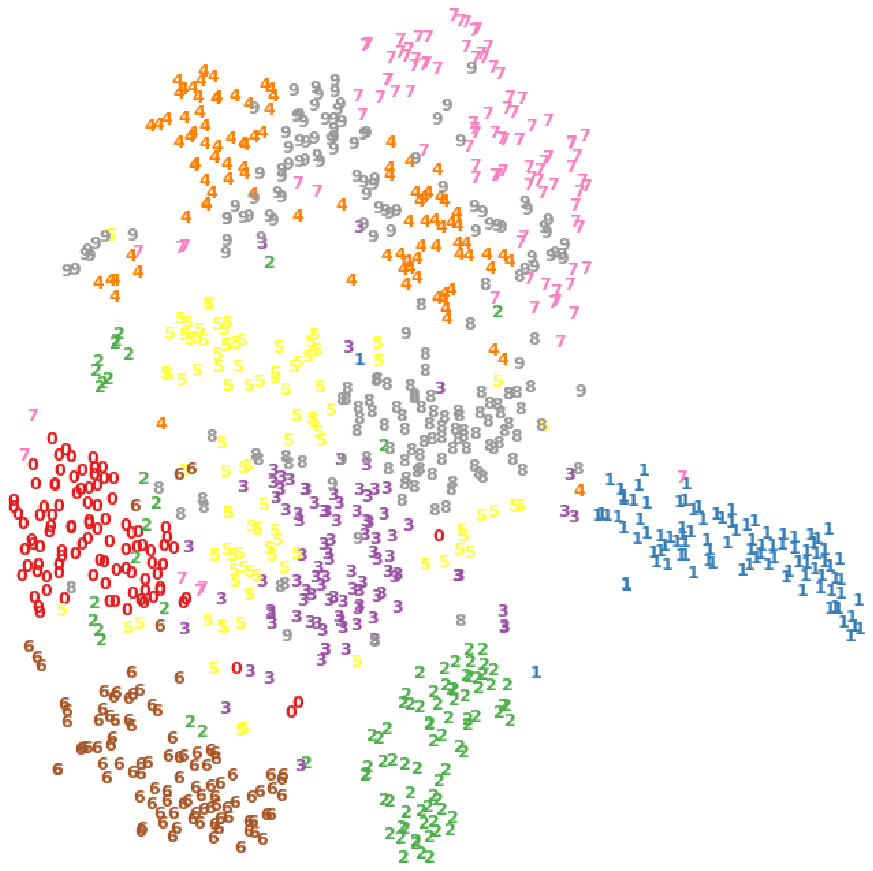}
  \caption{Original of $p$}
  \label{Original2.8.1}
\end{subfigure}
\begin{subfigure}{.5\textwidth}
  \centering
  \includegraphics[width=0.9\linewidth, height=0.8\textwidth, trim={20 20 20 20}, clip]{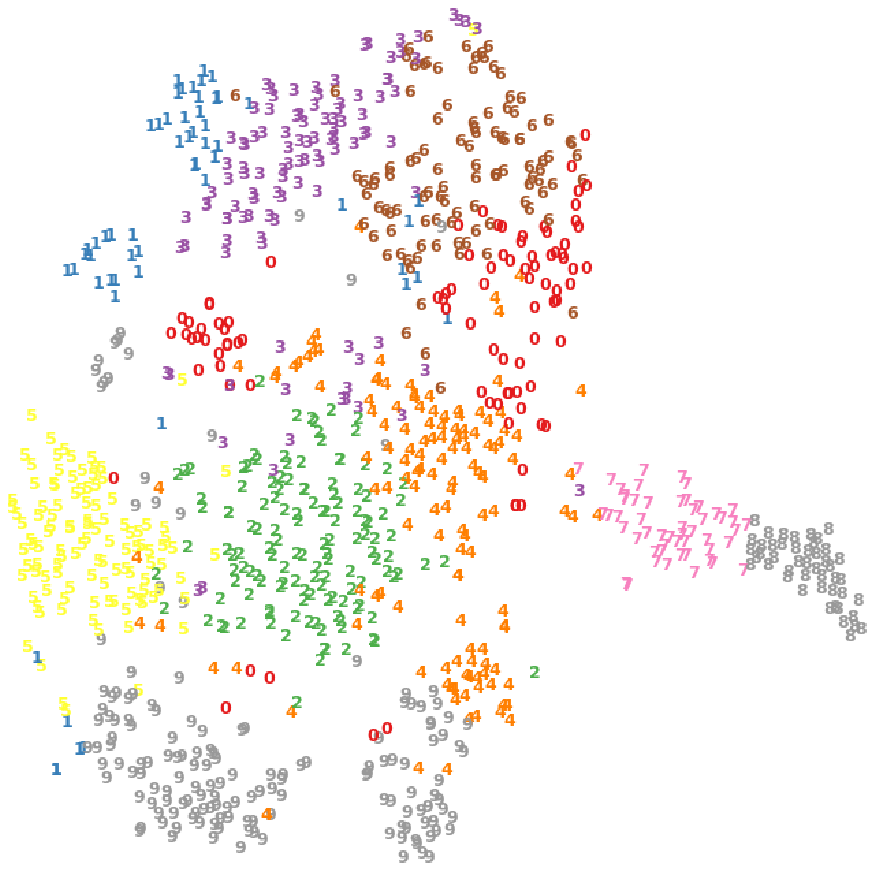} 
  \caption{Clustering of $p$}
  \label{Predicted2.8.1}
\end{subfigure}
\begin{subfigure}{.5\textwidth}
  \centering
\includegraphics[width=0.9\linewidth, height=0.8\textwidth, trim={20 20 20 20}, clip]{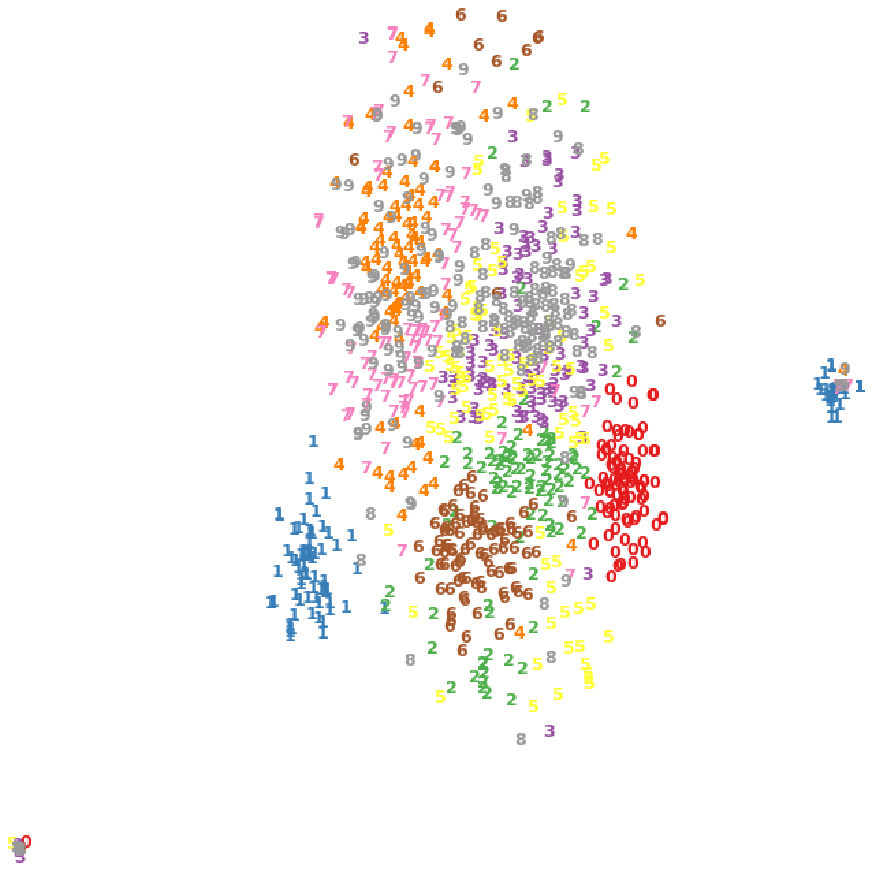}
  \caption{Original of $\theta$ and $\eta$}
  \label{Original2.8.2}
\end{subfigure}
\begin{subfigure}{.5\textwidth}
  \centering
\includegraphics[width=0.9\linewidth, height=0.8\textwidth, trim={20 20 20 20}, clip]{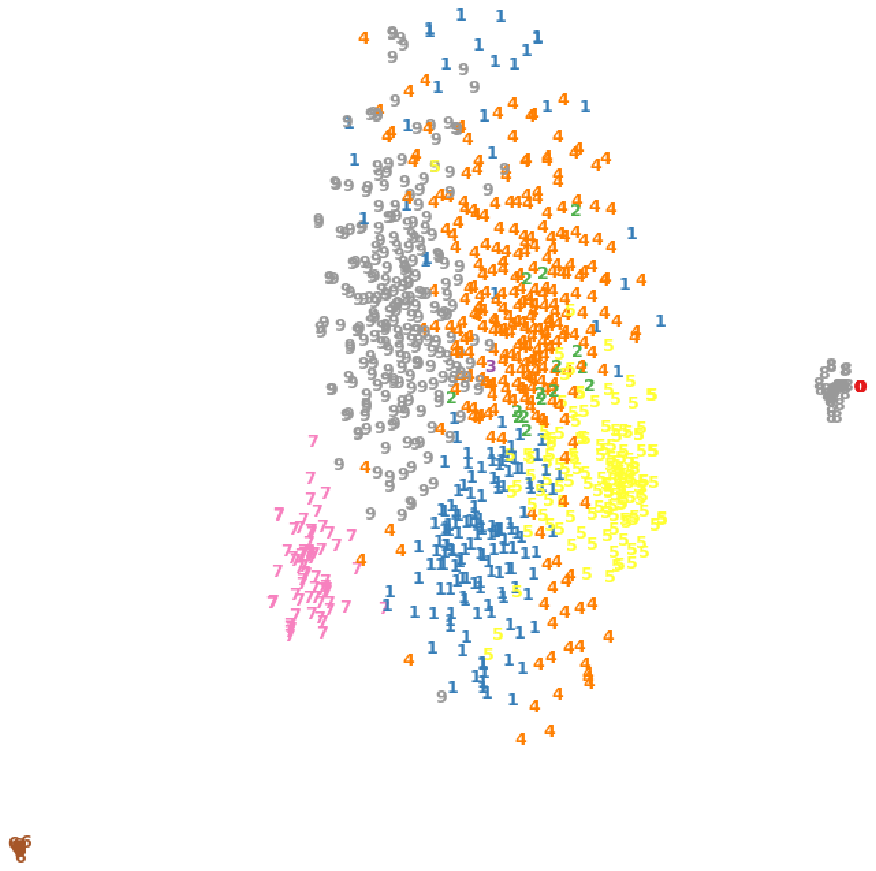}
  \caption{Clustering of $\theta$ and $\eta$}
  \label{Predicted2.8.2}
\end{subfigure}
\begin{subfigure}{.5\textwidth}
  \centering
\includegraphics[width=0.9\linewidth, height=0.8\textwidth, trim={20 20 20 20}, clip]{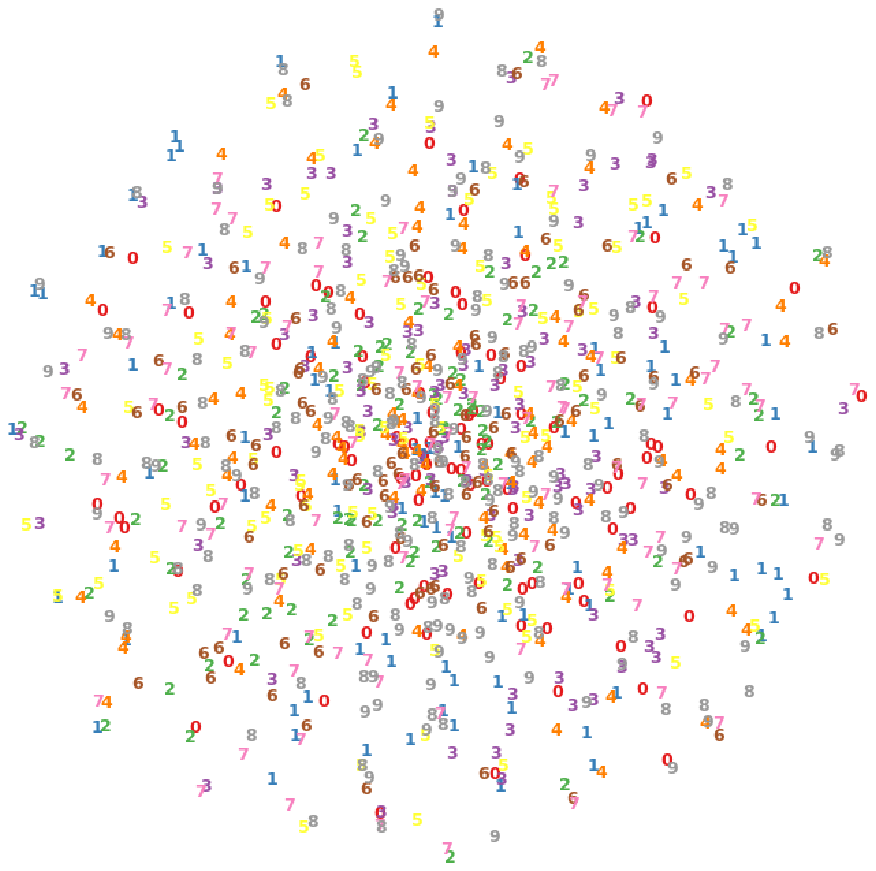}
   \caption{Original of $\theta$ and $\eta$ in $\beta$}
  \label{Original2.8.3}
\end{subfigure}
\begin{subfigure}{.5\textwidth}
  \centering
\includegraphics[width=0.9\linewidth, height=0.8\textwidth, trim={20 20 20 20}, clip]{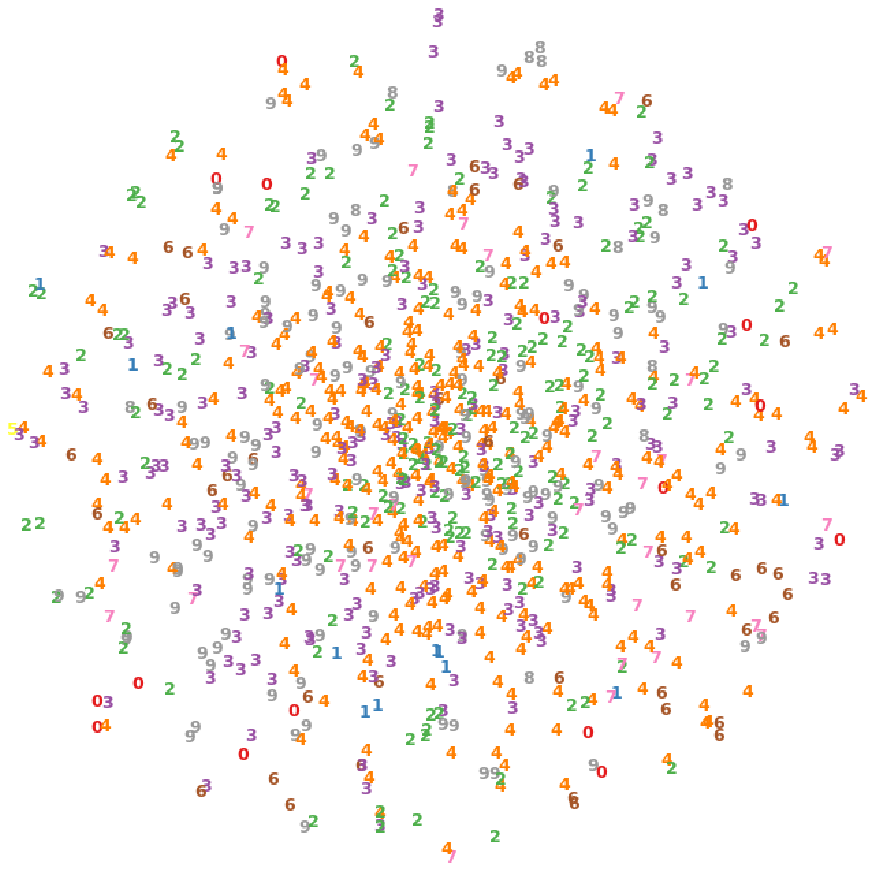}
  \caption{Clustering of $\theta$ and $\eta$ in $\beta$}
  \label{Predicted2.8.3}
\end{subfigure}
\end{figure}

By comparing the results of Table~\ref{res4.4.3} with Table~\ref{res4.4.2}, the clustering effect is improved to a certain extent. However, this doesn't mean that these parameters are suitable for low-rank representation tasks. Firstly, the AMI in Table~\ref{table13} reaches 0.5, and the parameter $p$ used can also be considered as the updated parameter in the submanifold, but the element value of the normalized reconstructed tensor has been saved in the $p$ of the last iteration which cannot be demonstrated as a special attribute in the submanifold. Secondly, although the AMI in Table~\ref{table14} also reached 0.46, the effect is improved more because the digits 0 and 6 have a good separation in the original distribution of parameters, which reduced the difficulty of clustering these digits to some extent. From Figure~\ref{Original2.8.2}, it is difficult to distinguish other digits, such as the digits 4,5 and 9 are almost mixed together. Therefore, the unfolded feature on single image tensor don’t own the property of low-rank representation.\\
\emph{Strangely, the results of Table~\ref{table15} is so much worse than Table~\ref{table14} that the parameter distribution in Figure~\ref{Original2.8.3} is almost impossible to use directly, which is not intuitive that parameters in $\beta$ should hold more key and effective information than the larger whole.}

 \section{Future Directions}
\subsection{Connecting with Neural Network}
$\beta$ plays a key role in the Legendre decomposition iteration. The $\theta$ in $\beta$ is used to compute $p$ (the reconstructed value), hence $\beta$ can be regarded as a dimension reducing choice of parameters, where key information locations are stored in theory. In each iteration, the $\beta$ and $p$ values are updated according to all the values of $\theta$. The number of updated parameters is relatively small, but the input tensor X is eventually restored from p.
$$q_{v}=\frac{1}{\exp (\psi(\theta))} \prod_{u \in \downarrow v} \exp \left(\theta_{u}\right), \quad \downarrow v=\{u \in B | u \leq v\}$$
The Legendre decomposition is more similar to the dropout mechanism if it's connected with the structure of the multi-layer perceptron. Dropout is often used in dense fully connected networks as a regularization method to reduce overfitting by preventing co-adaptation in training stage. For a neural network unit, dropout randomly invalidates a portion of hidden nodes and temporarily discards them from the network according to a certain probability. Therefore, we can use Legendre decomposition instead of dropout's random selection mechanism. Since dropout is used before each forward propagation, if a certain probability distribution is used to generate probability for each node before each propagation, Legendre decomposition can be used to reconstruct the tensor of probability distribution, and the core hidden layer node can also be selected as the update node of this activation, and the rest nodes can be temporarily deleted.

The other is intuitively connected with neural network compression. For the compression of convolutional layer and fully connected layer, the method of low-rank factorization can be used to achieve the effect of network compression acceleration. The CNN convolution kernel is a 4D tensor and the fully connected layer is a 2D tensor which may contain a lot of redundancy, so it is suitable to treat the network weight as a full-rank tensor and approximate the tensor with multiple low-rank tensors. Compared with CP decomposition and Tucker decomposition, Legendre decomposition may further reduce the time complexity, since the running time of Legendre decomposition is much less than other methods when the tensor size is small ($20^3\sim50^3$). It should be noted that $1 \times 1$ convolution cannot be achieved by tensor decomposition, and a great deal of retraining is required to achieve convergence.

In addition, the connection with boltzmann machine is inspired from Legendre decomposition\cite{sugiyama2018legendre}. By representing boltzmann machine as an undirected graph, the partial order relationship between nodes on the graph is established to form a hasse graph. Therefore, Legendre decomposition is a generalization of boltzmann machine learning in this way which aims to advance machine learning methods by leveraging tensor network representations. The design idea of Legendre decomposition may be more used in the exploration of better representation of spiking neural networks (SNN) since SNN unit fires only when the membrane potential reaches certain threshold which simulates the way synapses encode information through electrical impulses and is suitable for the representation of the discrete form of tensors.

\subsection{Connecting with Other Applications of Low-rank Representations}
\begin{figure}[!htbp]
\begin{subfigure}{.5\textwidth}
  \centering
  \includegraphics[width=0.9\linewidth, height=0.8\textwidth, trim={20 20 20 20}, clip]{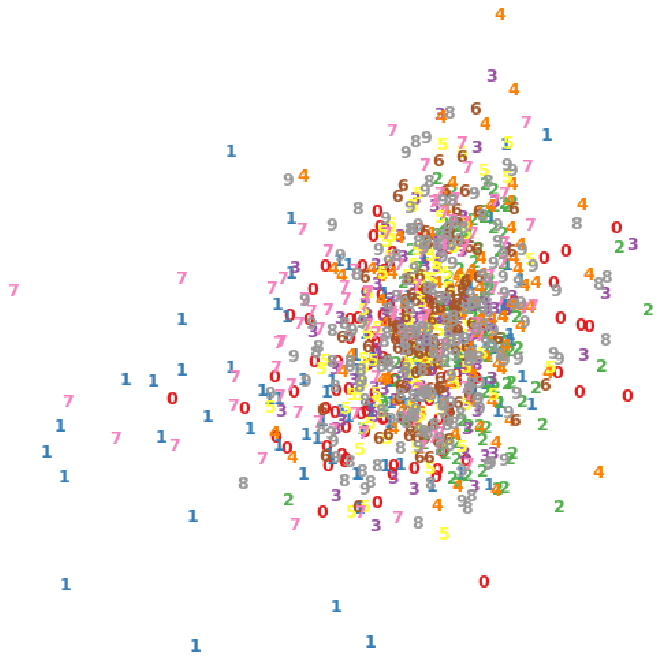}
  \caption{Original of $\sum{\theta}$ and $\sum{\eta}$}
  \label{Original2.7.1}
\end{subfigure}
\begin{subfigure}{.5\textwidth}
  \centering
  \includegraphics[width=0.9\linewidth, height=0.8\textwidth, trim={20 20 20 20}, clip]{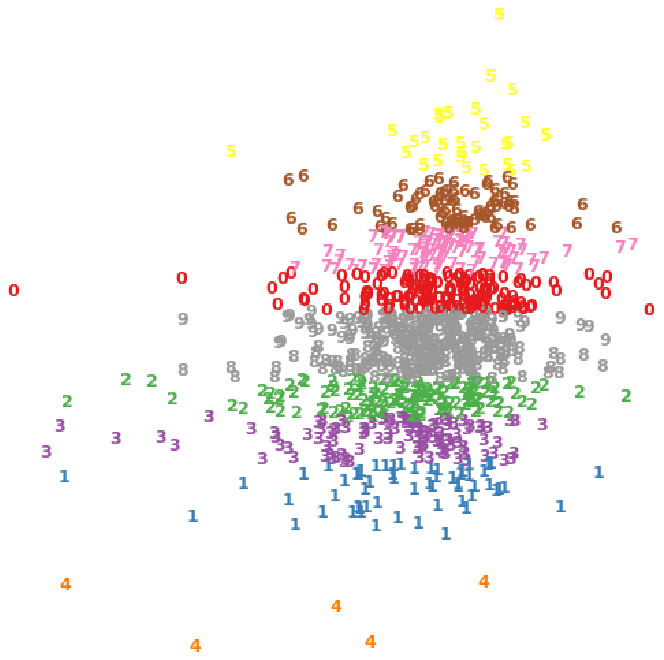} 
  \caption{Clustering of $\sum{\theta}$ and $\sum{\eta}$}
  \label{Predicted2.7.1}
\end{subfigure}
\begin{subfigure}{.5\textwidth}
  \centering
\includegraphics[width=0.9\linewidth, height=0.8\textwidth, trim={20 20 20 20}, clip]{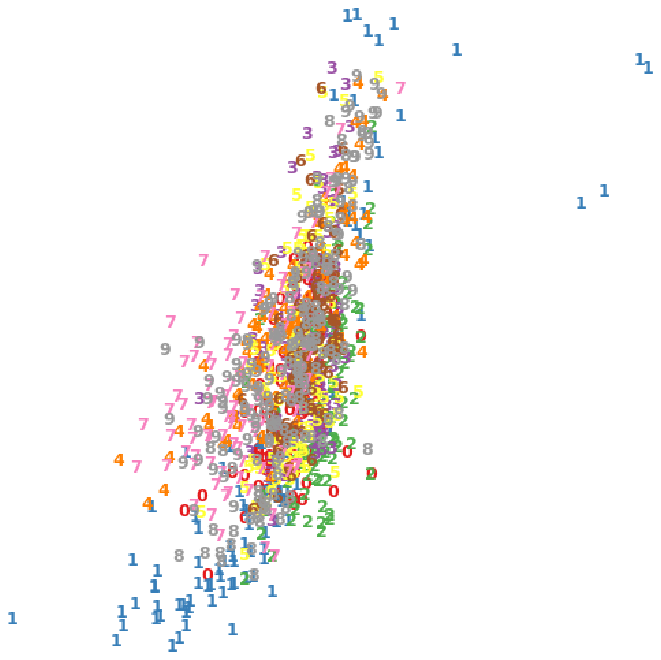}
  \caption{Original of $\sum{\theta}$ and $\sum{\eta}$ in $\beta$}
  \label{Original2.7.2}
\end{subfigure}
\begin{subfigure}{.5\textwidth}
  \centering
\includegraphics[width=0.9\linewidth, height=0.8\textwidth, trim={20 20 20 20}, clip]{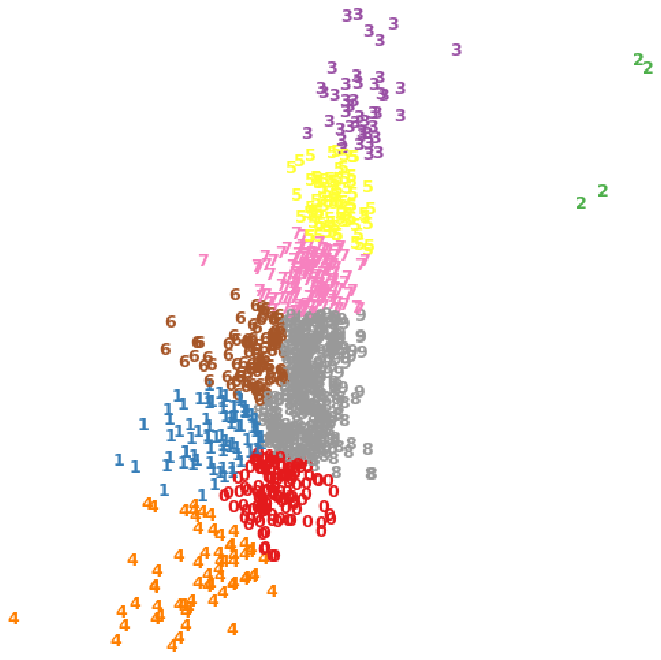}
  \caption{Clustering of $\sum{\theta}$ and $\sum{\eta}$ in $\beta$}
  \label{Predicted2.7.2}
\end{subfigure}
\begin{subfigure}{.5\textwidth}
  \centering
\includegraphics[width=0.9\linewidth, height=0.8\textwidth, trim={20 48 20 20}, clip]{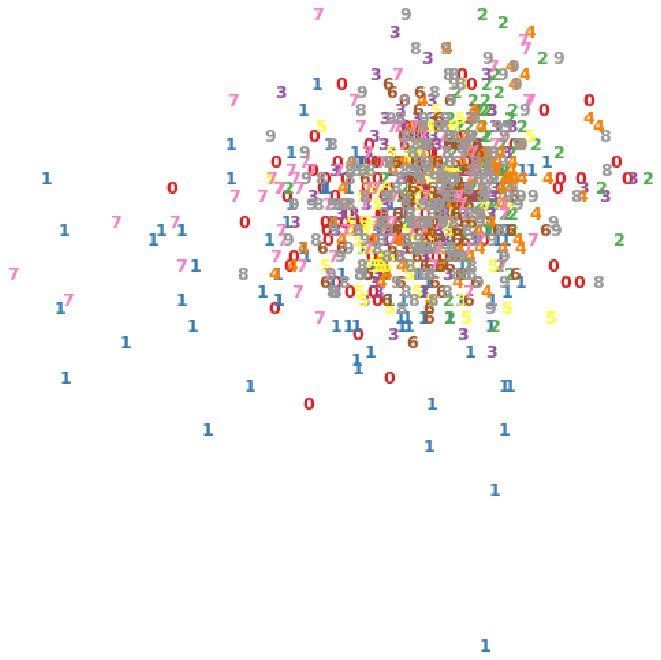}
  \caption{Original of $\sum{\theta}$ and $N_{noneZero}$}
  \label{Original2.7.3}
\end{subfigure}
\begin{subfigure}{.5\textwidth}
  \centering
\includegraphics[width=0.9\linewidth, height=0.8\textwidth, trim={20 48 20 20}, clip]{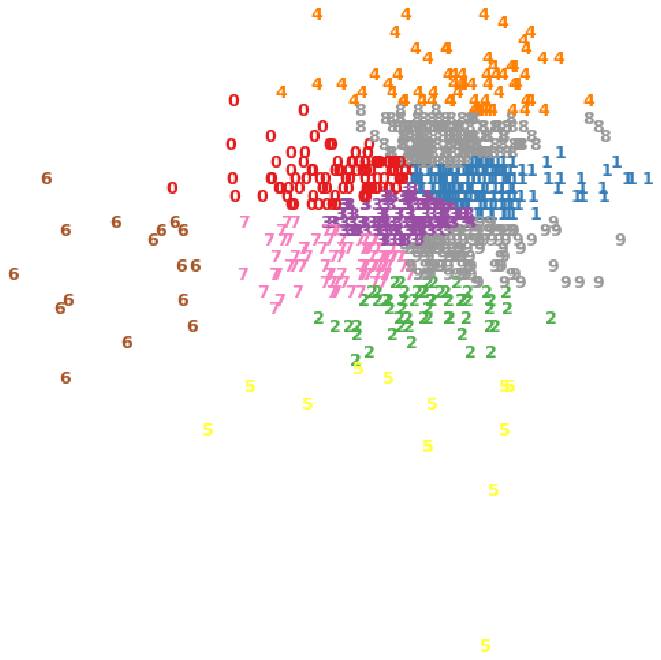}
  \caption{Clustering of $\sum{\theta}$ and $N_{noneZero}$}
  \label{Predicted2.7.3}
\end{subfigure}
\end{figure}

$\theta$ in $\beta$ can be computed to get reconstruction value $p$, there is only one exponential family relationship. Intuitively, the value of $\theta$ in $\beta$ contains information about the reconstructed value. However, from the perspective of discrete distribution visualization in the clustering experiment given in Figure~\ref{Predicted2.7.2} and Figure~\ref{Predicted2.8.3}, $\theta$ in $\beta$ did not reach the expectation that it might save the key information of the original input tensor from Section~\ref{Investigation}.

One of the hypotheses for the experimental results is that the dual parameters $(\theta,\eta)$ are used as the coordinate system on the submanifold, which is far different from the coordinate definition and meaning on Euclidean space, so the mapping relationship between the submanifold space and the standard 3D space coordinates cannot be established directly, and the information on the submanifold is more complex and difficult to use. Therefore, it may not be possible to directly use the information contained in the parameter as a low-dimensional representation of the tensor. Therefore, if we want to obtain an effective low-rank representation through Legendre decomposition, we need to find a special transformation method for the parameter or establish an isomorphism relation with the common manifolds.

If an effective rank reduction method can be found, Legendre decomposition can be used for other applications of low-rank representation. It's so easy to think about one of tasks is to predict short time series\cite{ghalamkari2020rank} which projects higher-order tensors to compressed core tensors by applying tensor decomposition. In addition, another application that utilizes tensor network to obtain better representations is multi-modal learning\cite{tensoremnlp17}. The features of single modes are regarded as low-rank subtensors on each dimension, and the final multi-modal representation is obtained through the fusion of subtensors. Each component of the whole tensor coordinate point corresponds to the information of different modes respectively. If Legendre decomposition is used as a means of tensor fusion, the representation may be more natural in the subspace compared to the 3-fold Cartesian space.

\section{Conclusions}
By introducing dual parameters in information geometry, the target tensor is transformed into multi-dimensional probability distribution in dually flat manifold, and global convergence is guaranteed by optimizing the KL divergence from the input tensor. Legendre decomposition combines the restructurable tensor with the information geometry, and obtains the unique decomposition of the given non-negative tensor by optimizing the parameters of the basis with partial order restriction in the submanifold. In addition, the form of dual coordinates conforms to the exponential family function, which can be combined with the probabilistic graphical model of boltzmann machine to some extent, which is very promising.\\
We analyze Legendre decomposition in theory and application. In terms of theory, we review the information geometry used and analyze the processes of tensor projection, and the dual parameters upward and downward computing. In terms of application, we have conducted clustering experiments with statistical characteristic based on image-batch tensors and unfolded feature based on single image tensor, aiming to find out parameters with low-rank representation ability. We analyze the results of the clustering experiments, which show that the parameters in the submanifold space are more complex and have no semantic information that can be used directly.\\
Therefore, due to the combination of Legendre decomposition with probabilistic model and the potential ability to select important parameters, we point out its future work: Firstly, the mapping of structural or utility connected with dropout and boltzmann machine, and the usage in the field of neural network compression as a new tensor decomposition method. Secondly, for the purpose of seeking low-rank representation in tensor decomposition, the combination of Legendre decomposition with applications such as time series analysis and multi-modal learning is much more valuable.

\section*{Acknowledgement}
This work was supported in part by U.S. National Institutes of Health (NIH) grant P41GM103712 and R01GM134020, U.S. National Science Foundation (NSF) grant DBI-1949629 and IIS-2007595.

\bibliographystyle{unsrt}  
\bibliography{main}  

\newpage
\section*{Appendix I: The Algorithm of Legendre Decomposition}
\begin{algorithm}[H]
\caption{Function Legendre Decomposition}
\begin{algorithmic}[1]
    \STATE{$preprocess(X)$} \tcp*{Convert the input tensor X to P}
    \STATE{$makePosetTensor(X, S); prepareForBeta(S)$} \tcp*{build S from X}
    \STATE{$makeBetaCore(S, beta, core\_size, false)$} \tcp*{takes partial order on S and initialize beta from given core\_size}
    \STATE{$initialize(S)$} \tcp*{initialize the S matrix}
    \REPEAT
        \STATE{$eProject$ OR $grad$} \tcp*{natural gradient or gradient descent}
        \STATE{$computeResidual$}
        \IF{$residual\_prev >= EPSILON$ AND $residual > residual\_prev$}
            \RETURN $step$
        \ENDIF
    \UNTIL{$step\geq repeat\_max$}
\end{algorithmic}
\end{algorithm}

\begin{itemize}
\item[Step 1.] Input a large number of parameters and carry out a transpose, and the original $(i,j,k)$ become $(k,i,j)$ in the implementation.
\item[Step 2.] Select the approximation mode, then enter the header file for Legendre decomposition.
\item[Step 3.] Initialize the original matrix, sum all elements in $X$, and divide by sum for normalization.
\item[Step 4.] Initialize node matrix $S$. \\
$S.p$ is the recorded value within the node, which is used to reconstruct; \\
$S.theta$ is $log(p)$, which is used to compute;\\
$S.Nonezero$ records whether P is non-zero. \\
At the same time, S(0,0,0) is initialized as non-zero and not as a basis.
\item[Step 5.] For the initial S node matrix in step 4, calculate the node value (calculate from the maximum end $(i,j,k)$ to $(0,0,0)$ during the calculation).
$$\eta_{v}=\sum_{u \in \uparrow v} q_{u}=\sum_{u \in \Omega} \zeta(v, u) q_{u}, \quad \uparrow v=\{u \in \Omega | u \geq v\}$$
It should be emphasized here that the e-projection is selected in this paper, because it generally takes fewer parameters (the decomposition basis is generally a smaller subset). In e-projection, the above equation needs to be modified to directly replace $q_u$ with $p_u$.
\item[Step 6.] Prepare for $\beta$, the base selection operation of the decomposition basis (in initialization, $(1,1,...,1)$ is not used as the decomposition basis).
$$\Omega^{+}=\Omega \backslash\{(1,1, \ldots, 1)\}$$
P\_tmp is a mark symbol, which represents whether it is in the decomposition basis. Here, there are two modes. The simple mode selects the minimum core\_size non-zero value as the decomposition basis, while the complex mode chooses the decomposition basis by jumping according to the core\_size as the interval.
\item[Step 7.] Initialize $\eta$ in $S$. All the real $\eta$ values ($\hat{\eta}_{v}$) were calculated beforehand. In addition, the coordinates of $B$ were selected and stored in beta.second, which has not changed. 
$$\mathcal{S}_{\mathcal{P}}=\left\{\mathcal{Q} \in \mathcal{S} | \eta_{v}=\hat{\eta}_{v} \text { for all } v \in A\right\}$$
Step 7 corresponds to original values $\hat{\eta}$, which is brought into the calculation during iteration.
\item[Step 8.] Iteration.\\
gradient descent or natural gradient;\\
compute $p$, As shown in Figure~\ref{rp};\\
renormalize, repeat step 3;\\
compute $\eta$, repeat step 5.
\end{itemize}
\newpage
\section*{Appendix II: 2D/t-SNE Visualizations with Clustering of Different Parameters}

\end{document}